\documentclass[fleqn,10pt]{olplainarticle}
\usepackage[obeyFinal]{easy-todo}
\usepackage{rotating}
\usepackage{algorithm} 
\usepackage{algpseudocode}
\usepackage{textcomp}
\usepackage{url}
\usepackage{multirow}
\usepackage{diagbox}
\usepackage[T1]{fontenc}

\author[1]{Krystian Strza\l{}ka}
\author[1]{Szymon Mazurek}
\author[2]{Maciej Wielgosz}
\author[2]{Pawe\l{} Russek}
\author[1]{Jakub Caputa}
\author[1]{Daria \L{}ukasik}
\author[1]{Jan Krupi\'nski}
\author[1]{Jakub Grzeszczyk}
\author[1]{Micha\l{} Karwatowski}
\author[1]{Rafa\l{} Fraczek}
\author[2]{Ernest Jamro}
\author[1]{Marcin Pietro\'n}
\author[2]{Sebastian Koryciak}
\author[2]{Agnieszka Dabrowska-Boruch}
\author[2]{Kazimierz Wiatr}

\affil[1]{ACK Cyfronet AGH, ul. Nawojki 11, 30-950 Krakow, POLAND}
\affil[2]{AGH University of Science and Technology, 30 Mickiewicza Ave., 30-059 Krakow, POLAND}

\title{Using simulation to calibrate real data acquisition in veterinary medicine }


\begin{abstract}
This paper explores the innovative use of simulation environments to enhance data acquisition and diagnostics in veterinary medicine, focusing specifically on gait analysis in dogs. The study harnesses the power of Blender and the Blenderproc library to generate synthetic datasets that reflect diverse anatomical, environmental, and behavioral conditions. The generated data, represented in graph form and standardized for optimal analysis, is utilized to train machine learning algorithms for identifying normal and abnormal gaits. Two distinct datasets with varying degrees of camera angle granularity are created to further investigate the influence of camera perspective on model accuracy. Preliminary results suggest that this simulation-based approach holds promise for advancing veterinary diagnostics by enabling more precise data acquisition and more effective machine learning models. By integrating synthetic and real-world patient data, the study lays a robust foundation for improving overall effectiveness and efficiency in veterinary medicine.

\end{abstract}

\begin{document}

\maketitle

\flushbottom

\thispagestyle{empty}

\section{Introduction}

As we advance further into the digital age, the veterinary field is no exception to the rapid technological developments taking place. This evolution has led to a progressive expansion in data availability, both in terms of volume and diversity. With data playing an increasingly central role in the delivery of veterinary care, optimizing the process of data acquisition is a critical goal. Data must not only be accurate, but it should also be contextually relevant and easy to interpret by the veterinary professionals to enable effective decision-making. Nevertheless, ensuring the accuracy and reliability of real-world data acquisition can often be a challenge, considering the variability of animal conditions, environmental factors, and limitations in existing technology.

Emerging technologies such as advanced computer simulation, machine learning, and synthetic datasets offer transformative opportunities to overcome these challenges. In fact, the application of these technologies has already made a significant impact in various medical fields, showing promising results in enhancing diagnosis, treatment, and overall patient care. By creating a virtual environment that closely mimics the real world, simulation technologies can help veterinary professionals to acquire, interpret, and utilize data more effectively.

This paper specifically focuses on the application of computer simulation in calibrating real data acquisition in veterinary medicine. We delve into how such a method can be utilized to fine-tune data acquisition techniques, improve diagnostic accuracy, and develop more robust machine learning algorithms.

We hypothesize that the integration of computer simulation with real-world data acquisition will not only enhance the accuracy and reliability of veterinary data, but also lead to improved patient outcomes. This hypothesis stems from the unique capabilities of computer simulation to replicate various aspects of the real world, including animal anatomy, behavior, and environmental factors.

The ensuing sections discuss how such an innovative approach can be applied in veterinary medicine, and how it could potentially revolutionize the field by enhancing the quality of data acquisition and consequently, the standard of veterinary care.

Utilizing simulation to calibrate real data acquisition in veterinary medicine is an innovative approach that can greatly enhance the quality and accuracy of diagnostic and therapeutic procedures. By incorporating advanced computer graphics, biomechanical models, and synthetic datasets, veterinary professionals can fine-tune their data acquisition techniques and equipment settings to obtain optimal results in real-world scenarios. These simulations can closely replicate the diverse range of animal anatomies, movements, and behaviours observed in practice, enabling the development of more precise and targeted diagnostic imaging and treatment protocols. In addition, by simulating various environmental factors such as lighting conditions, background noise, and distractions, veterinary practitioners can better understand the challenges and limitations of real-world data acquisition and adapt their techniques accordingly. The integration of simulated data with actual patient data can also facilitate the development of more robust and generalizable machine learning algorithms, ultimately improving the overall effectiveness and efficiency of veterinary medicine. By leveraging the power of simulation to calibrate real data acquisition, the veterinary field can continue to advance its understanding of animal health and wellbeing, paving the way for enhanced patient care and outcomes.

\section{Related work}
To the best of our knowledge, the was no other work tackling directly the problem that is the core contribution of this paper. However, the problem can be transferred to the domain of gait recognition (GR) using deep learning, where information describing the performed activity has to be extracted from spatial and temporal patterns present in the input data.
GR with deep learning is widely explored in humans (\cite{satnos2022gaitsurvey}). Common approaches exploit spatial features of the input data, using convolutional neural networks to extract the features allowing for gait classification. These approaches showed high performance on common GR datasets in humans, but also extremely high sensitivity to the context of the data (\cite{sokolova2017cnngaitrecognition}). Recurrent neural networks were also used to exploit temporal relationships in the data, reaching state-of-the-art on some benchmarks (\cite{tran2021lstmgaitrecoginition}). These networks usually use handcrafted features or work in combination with auto encoders that provide features learned from input data \cite{zhang2019gait}.
As can be seen in \cite{feng2022graphgaitsurvey}, graph based approaches using joint position are also an approach that enjoy high recognition. These solutions revolve mainly around exploiting spatiotemporal properties of the poses represented as a graph in time. This kind of analysis was mainly catalyzed by milestone works such as \cite{sijie2017stgcn} introducing spatiotemporal convolutions on graphs or \cite{shi2019agcn2s}, incorporating the attention mechanism. The growing popularity and high performance (\cite{xu2020attention, ding2019attention, cheng2020decoupling}) of approaches based on these methods indicate that they are suitable for performing the tasks of gait analysis.

\section{In-Depth Description of the Simulation Environment}
\subsection{Overview}
The simulation environment in question was meticulously designed using the powerful software Blender in conjunction with the Blenderproc library. The set-up allows the creation of random scenes featuring a loaded dog armature which can be observed and recorded from various angles of view. The output from each simulation run consists of an RGB video and a segmentation video. The latter overlays the dog mesh with a binary mask. Additionally, essential data related to pose, visibility of each joint, and a categorical label indicating the state of the dog's gait (healthy or unhealthy) are also saved. \
A single simulation run involves a systematic process, which includes:
\begin{enumerate}
\itemsep0em
\item Configuration of the renderer.
\item Set-up of the scene's lighting (incorporating light sources and HDRi).
\item Loading of the mesh and initiating animation.
\item Arrangement and set-up of objects within the scene.
\item Animation of movement.
\item Generation of RGB and Segmap outputs.
\item Saving pose data and visibility.
\end{enumerate}
\subsection{Renderer Configuration}
The renderer configuration is a key part of our simulation environment and is governed by a set of parameters defined in a config.yml file. We utilize the Cycles rendering engine, a highly advanced rendering engine incorporated in Blender, with a modest sample count (4-8) supplemented by the "OPTIX" denoiser. Given the low optical complexity of our rendered scenes, such a sample count, along with the applied denoising, suffices to generate satisfactory results.

\subsection{Lighting Setup of the Scene}
    The lighting setup within the scene plays an integral part in the realism of the simulation. This involves strategic placement of static light sources across the scene, accompanied by additional lighting generated using a random HDRi, part of the Blenderproc collection acquired from polyhaven.com.

\subsection{Mesh Loading and Animation Initiation}
    At this stage, the rigged armature for the dog model is loaded from a .blend file. The positions and rotations for each bone structure are loaded frame-by-frame, ensuring realistic motion. Location and rotation keyframes are then created for each individual bone and frame to maintain continuity and fluidity in movement.

\subsection{Scene Object Configuration}
    Setting up the objects within the scene involves several nuanced tasks. Initially, a floor is created with a randomized texture over a specified area. Objects on this floor are then introduced based on a predefined density value per square meter. Only objects with bounding volumes less than 10 and greater than .1 are selected for a balanced and realistic scene. As a safeguard against potential obstructions, objects that may impede the dog's path are carefully removed to avoid unrealistic collisions.

\subsection{Animating Movement}
    To ensure optimal data collection, the camera remains stationary throughout the simulation, keeping the dog model at the center of the scene. Rather than moving the dog itself, all other objects in the scene are animated to move around it, creating the illusion of the dog's movement while ensuring that the pose data is consistently centered in the scene.

    \subsection{Writing RGB + Segmap}
\begin{figure}[H]
            \centering
            \includegraphics[width=.45\textwidth]{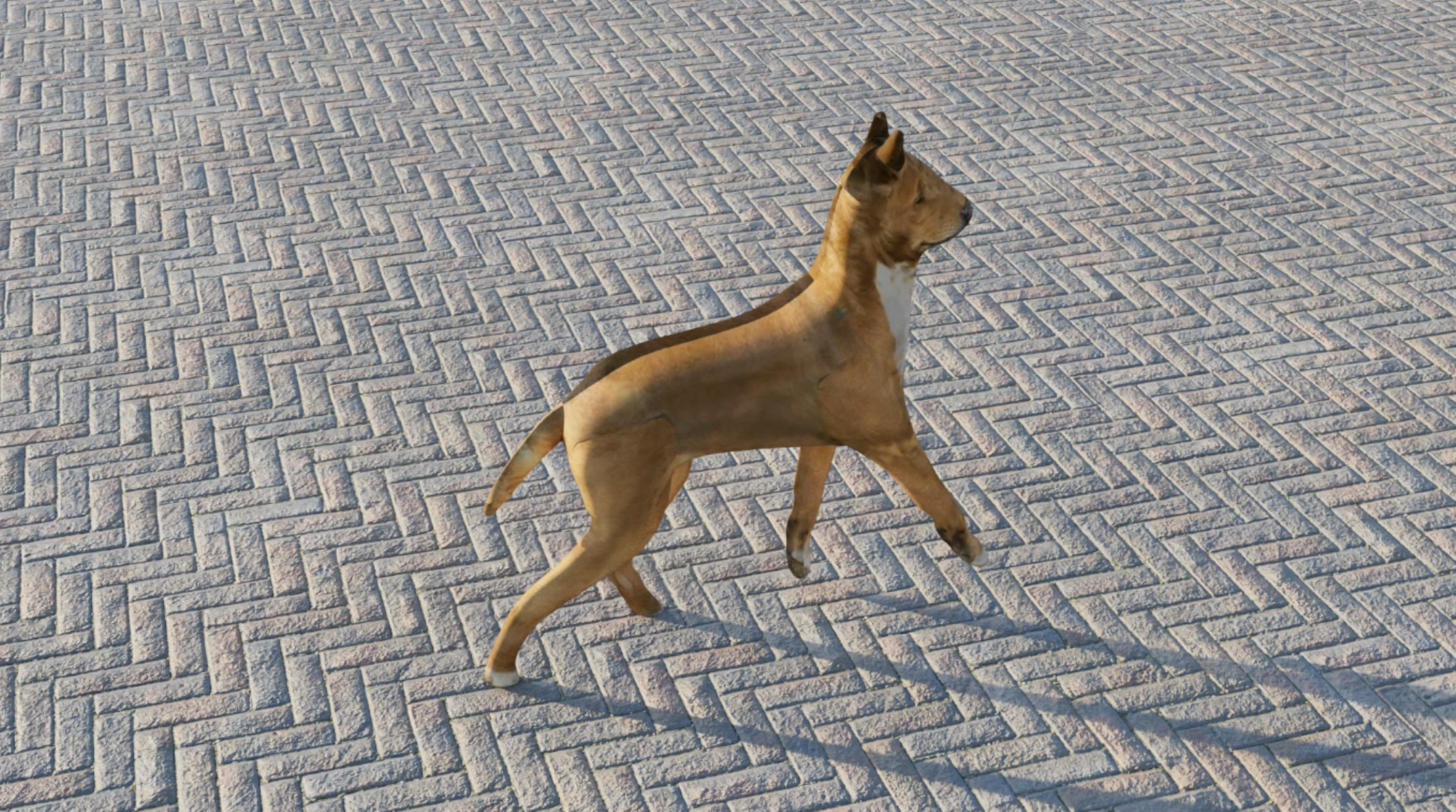}
            \includegraphics[width=.45\textwidth]{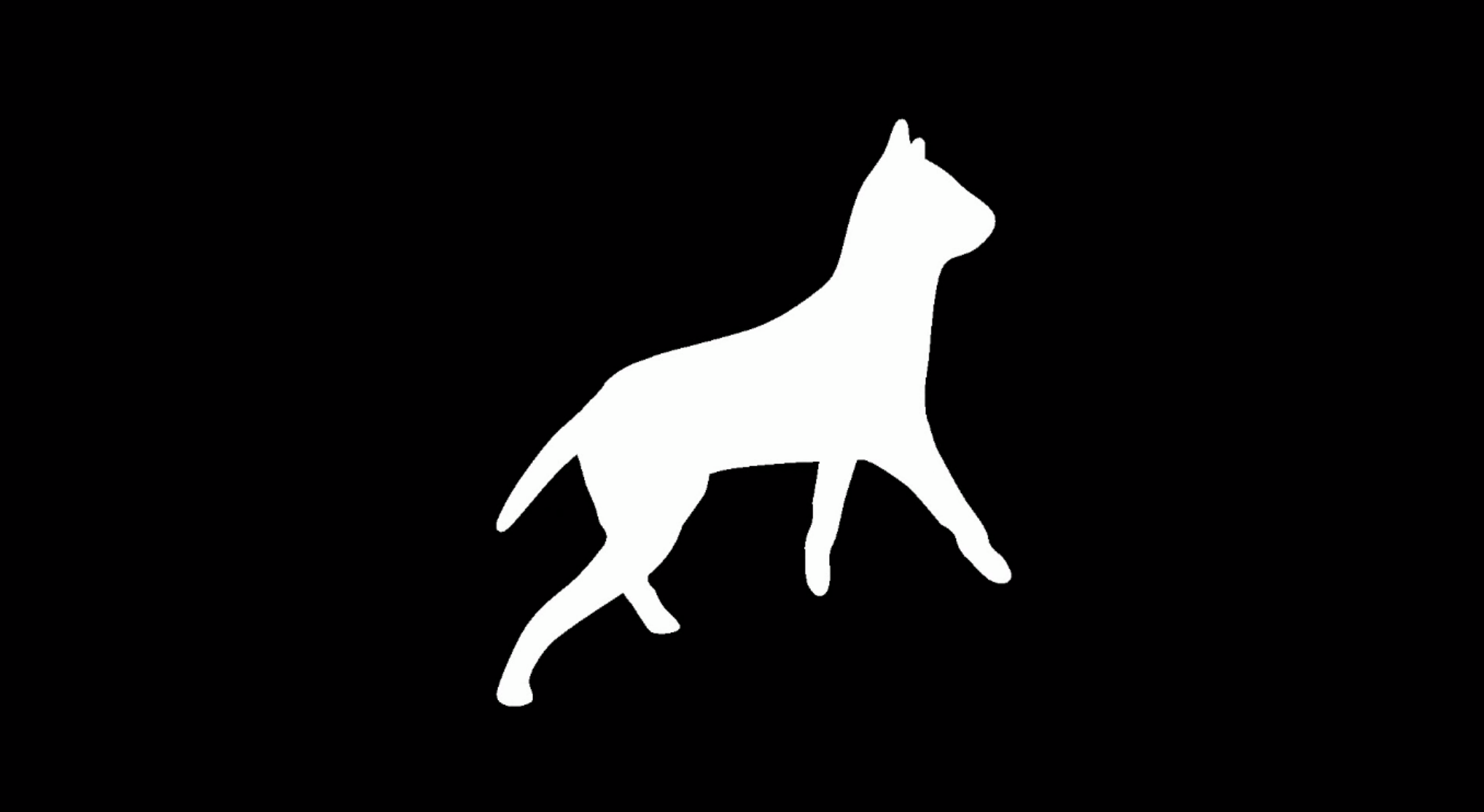}
            \caption{Example frame from simulator (left) and binary segmentation map (right)}
            \label{fig:segmented-rgb}
        \end{figure}

        The rendered RGB frames and the segmentation of the dog are encoded into an mp4 video format using ffmpeg. The segmentation data is represented as a binary image, with the dog mesh assigned a value of 255 and the background assigned a value of 0, as shown in Fig.\ref{fig:segmented-rgb}.
    
    \subsection{Saving Pose + Visibility}
\begin{figure}[H]
            \centering
            \includegraphics[width=\textwidth]{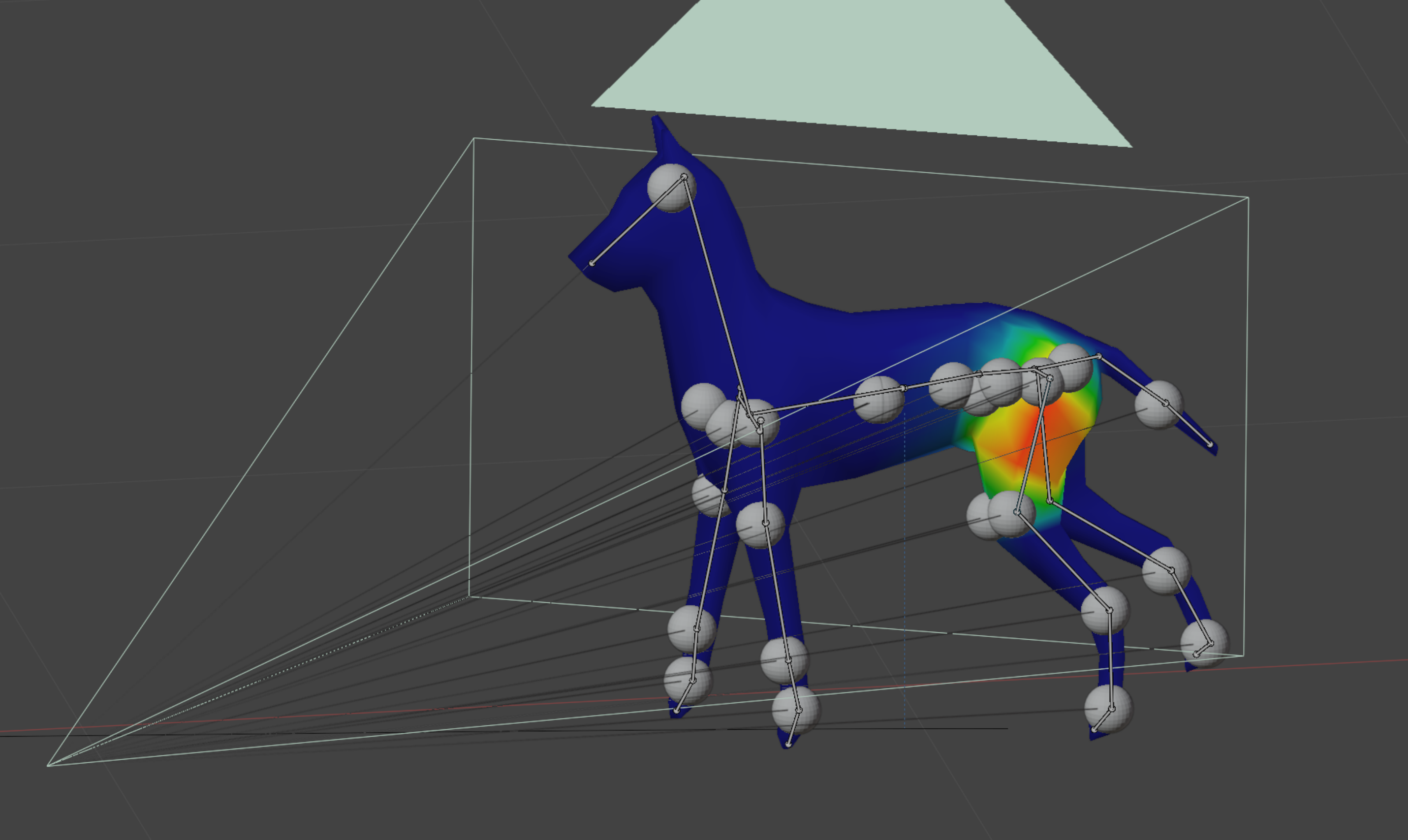}
            \caption{Raycasting visualisation with a heatmap of vertex groups which belongs to the rear, left hip. Lines indicate rays cast from the camera to each joint and spheres indicate points hit by the rays.}
            \label{fig:raycasting}
        \end{figure}

        The pose of the dog is saved in global coordinates to a JSON file. The saved pose includes the following information:
        \begin{enumerate}
            \itemsep0em
            \item Bone head location
            \item Bone tail location
            \item Bone head visibility
            \item Bone tail visibility
        \end{enumerate}
        Joint visibility is determined by casting a ray from the camera towards the position of each tested joint. If the ray intersects the face belonging to the joint's vertex group, the joint is marked as visible. If the ray hits any other object, the joint is marked as obstructed. Visualization of this approach can be seen Fig. \ref{fig:raycasting}. 
\section{Data preparation and preprocessing}
For the purpose of this study, we used the simulation environment to generate two major datasets with different granularity in terms of the ranges of camera angles used in the generated videos. The angle of the camera in a video was chosen randomly from pre-determined lower and upper boundaries. For the first dataset, the angle interval was 90, for the second one, it was set to 45. Each sub-dataset contained 15 videos for both abnormal and normal gait classes. The videos were 7 seconds long, with 25 frames per second. The structure of generated datasets is shown in Tab.{\ref{tab1-dataset}} and Tab.{\ref{tab2-dataset}}.

\begin{table}[H]
\centering
\begin{tabular}{l|l|l|l|l}
\begin{tabular}[c]{@{}l@{}}Dataset \\ number\end{tabular} & 1      & 2        & 3         & 4         \\ \hline
\begin{tabular}[c]{@{}l@{}}Angle \\ range\end{tabular}    & 0 - 90 & 90 - 180 & 180 - 270 & 270 - 360
\end{tabular}
\caption{Angle intervals of the sub datasets in the first dataset.}
\label{tab1-dataset}
\end{table}

\begin{table}[H]
\begin{tabular}{l|l|l|l|l|l|l|l|l}
\begin{tabular}[c]{@{}l@{}}Dataset \\ number\end{tabular} & 1      & 2       & 3        & 4         & 5         & 6         & 7         & 8         \\ \hline
\begin{tabular}[c]{@{}l@{}}Angle \\ range\end{tabular}    & 0 - 45 & 45 - 90 & 90 - 135 & 135 - 180 & 180 - 225 & 225 - 270 & 270 - 315 & 315 - 360
\end{tabular}
\caption{Angle intervals of the sub datasets in the second dataset.}
\label{tab2-dataset}
\end{table}

These datasets present an interesting opportunity to study the influence of the camera angle on the accuracy of our machine learning model. By varying the camera angles in the videos, we are adding a level of complexity to the dataset that may influence the recognition of abnormal and normal gait classes. Additionally, with the second dataset offering a finer granularity of camera angles, we can examine how this increased detail affects the model's ability to accurately classify the gait. We also acknowledge the possibility that different angles might provide different amounts of information, a factor that could potentially affect the performance of the model.

\subsection{Data Representation and Feature Preprocessing}
The raw data extracted from each video was organized in the form of a graph. In this graph representation, each joint in the simulated dog model was depicted as a node, thereby simplifying the intricate biological structures into an abstract yet informative format. 

To capture the genuine anatomical interconnections of the joints, an adjacency matrix was constructed. This matrix mapped the relationships between the nodes, preserving the underlying structure and interactions of the actual joint system.

Each node, representing a joint, was further associated with a feature matrix. This matrix was designed to capture the dynamics of each joint by recording the coordinates of the joint across consecutive frames. However, it was also necessary to account for instances where a joint was not visible in a given frame. In these cases, we introduced a masking value, -1, to substitute for the coordinates. This masking value served as a placeholder and a signal for the absence of data.

The next step in the process involved standardizing the features. Each coordinate vector associated with a node was z-standardized across the time dimension, inspired by the approach taken by \cite{aderinola2022gaitbased}. This procedure was executed to normalize the data and mitigate the influence of outliers, thereby making the data more conducive to analysis. However, care was taken to ensure that masked coordinates, indicative of invisibility, were not involved in the calculation of mean or standard deviation and were left unaltered. This step was vital to maintain the integrity of the masking system and to avoid skewing the standardized results. 

Our approach to data representation and feature preprocessing is centered around preserving the real-world complexity and dynamics of the joint system, while enabling effective computational modeling and analysis. This forms the foundation for our subsequent machine learning work.

\section{Classification}

\subsection{Experimental approach}
To assess the performance on every camera settings on the classifier's performance, stratified k-fold cross validation was performed on every dataset to assign videos into training or testing subsets. The training data was further split to create additional validation set, containing 20\% of the videos in the original training subset. The number of folds was set to 5, with stratification based on the video labels present in the dataset. The influence of the sample length on classification performance was also taken into account via evaluating samples of 5, 10, 15 and 30 frames with 50\% overlap.
For every fold and sample length, a separate model was trained and evaluated. To allow for reproducibility and fair comparison between folds and datasets, every data split as well as model initialization was done with the same random seed. Algorithm \ref{algorithm1-procedure} describes the procedure.

\begin{algorithm}
    \begin{algorithmic}[1]
        \For {All angle interval groups}
        \State Prepare list of videos with their class labels
        \For {All evaluated timestep lengths}
        \State Perform stratified k-fold split of the videos into training and testing group
        \For {Every fold}
        \State Load the data
        \State Extract samples with chosen timestep length and overlap
        \State Partition data into training, validation and testing subsets
        \State Initialize and train the model
        \State Evaluate the model on the test subset for given fold
        \EndFor
        \State Calculate mean result of every metric across all folds
        \EndFor
        \EndFor
    \end{algorithmic}
    \caption{Algorithm describing the experimental approach.}
    \label{algorithm1-procedure}
\end{algorithm}

\subsection{Graph Neural Networks}
Graph Neural Networks (GNNs) are a class of learning models designed to handle data structured as graphs. Unlike traditional neural networks, which excel in processing grid-like data (e.g., images, sequential data), GNNs leverage the inherent relationships and connections within graph data for learning and prediction. A GNN operates by propagating information across nodes of a graph, enabling the capture of complex patterns among node features and relationships. The iterative message-passing scheme inherent to GNNs allows them to integrate local and global information from a graph efficiently. This characteristic renders them effective tools in diverse domains, including social network analysis, molecular chemistry, and transportation networks, among others. Despite their capabilities, GNNs face challenges, such as handling dynamic graphs and scaling with large graph datasets, which are active research areas.

\subsection{Application of Graph Neural Networks to Canine Simulations}
The incorporation of Graph Neural Networks (GNNs) in canine behavior simulations can be a significant enhancement in our methodology. The power of GNNs lies in their ability to capture and model complex relationships and dependencies, precisely the kind of interactions that define a group of joins. Instead of considering each joint as an independent entity, GNNs allow us to model the entire population as an interconnected graph, where nodes represent individual joint and edges represent various interactions and relationships among them. Furthermore, GNNs have an inherent capacity to deal with variable-sized inputs, making them highly adaptable to different population sizes and dynamics. By leveraging these strengths of GNNs, we can enhance the accuracy and realism of our simulations.

\subsection{The Superiority of the A3T-GCN Model}
The A3T-GCN (Attention-based Adaptive Adjacency Matrix Time Graph Convolutional Network) model has emerged as a innovation in the field of spatiotemporal data prediction, demonstrating considerable superiority over other models. This model is particularly designed for spatiotemporal phenomena. By incorporating an attention mechanism, A3T-GCN dynamically adjusts the adjacency matrix, ensuring a more accurate prediction. This unique feature allows it to capture global variations efficiently, setting it apart from other models. Moreover, it demonstrates strong long-term prediction capability, maintaining high accuracy for prediction horizons ranging long in time. Its performance, even under varying time series lengths, remains stable, confirming its applicability in both short-term and long-term traffic forecasting tasks. The A3T-GCN model is thus a compelling choice for spatiotemporal data streams as flow of dogs joins in time.

\subsection{Classification model}
\begin{figure}
    \centering
    \includegraphics[scale=.15, keepaspectratio]{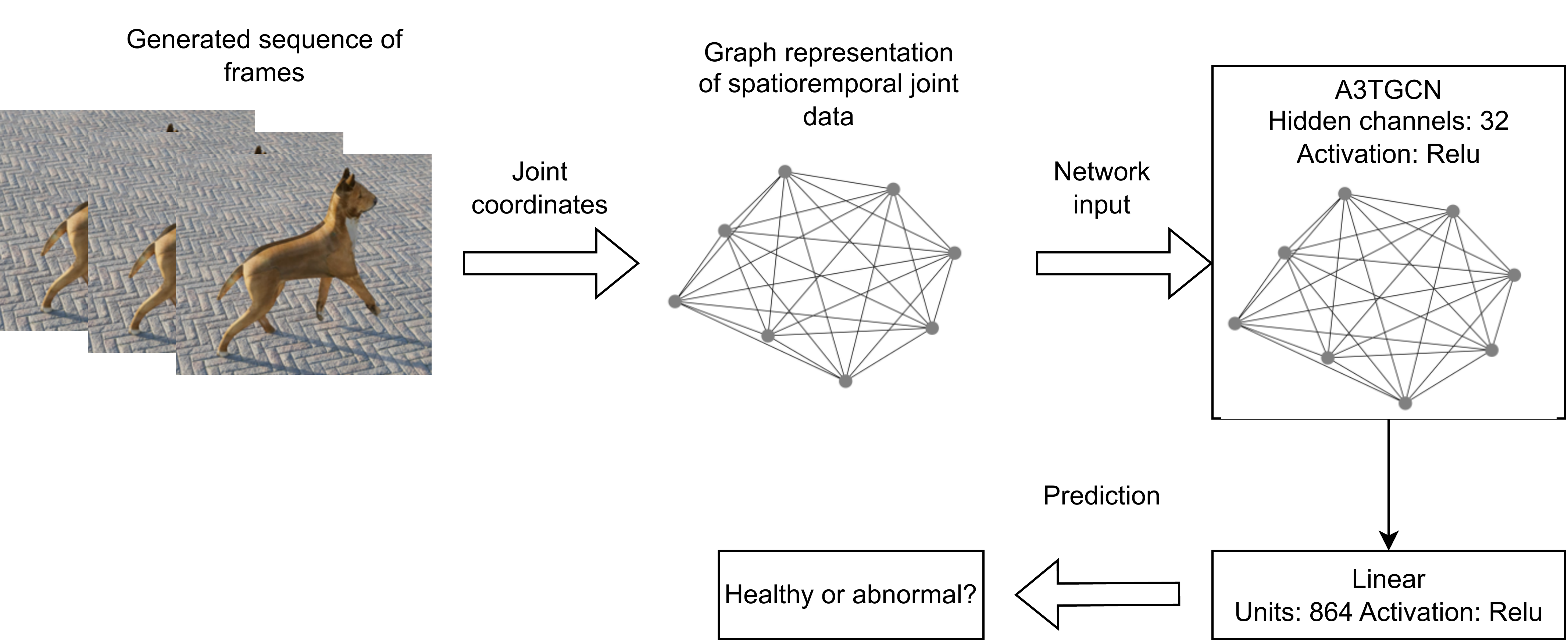}
    \caption{Proposed network architecture and data flow.}
    \label{fig:network}
\end{figure}

For the purpose of the experiments, simple temporal graph neural network model was created, based on AT3GCN graph temporal convolutional layer \cite{zhu2020a3tgcn}. The architecture and hyperparameters of the model can be seen in the Fig.\ref{fig:network}. No additional parameter optimization nor architecture search was performed, as the main goal of the study was to obtain information about the preferred recording setup for preparation of the input videos for different sample lengths.

\subsection{Implementation details}
The experiments were conducted on Athena supercomputer, using two AMD EPYC 7742 64-Core CPUs, 20 GB RAM and single Nvidia A100 GPU. The code was created with Python 3.1.9 using Pytorch 1.13.1 \cite{paszke2019pytorch} with Lightning 2..2 \cite{falcon2019lightning} wrapper and additional Pytorch Geometric 2.2.0 \cite{fey2019torchgeometric} and Pytorch Geometric Temporal .54.0 \cite{rozemberczki2021torchgeotemp} libraries for graph data processing.
The model was trained maximum 30 epochs using batch size of 8. Parameter optimization was performed using AdamW \cite{loshchilov2017adamw} optimizer with learning rate of .002 and weight decay of .01. Early stopping algorithm was used to prevent overfitting to the data, stopping the training if no progress on validation loss was noted for 6 consecutive epochs. The model displaying the lowest validation loss was used for final evaluation after training.

\section{Results}

Initially, we carried out a series of experiments focusing on datasets that encompassed a broad spectrum of angles. These datasets were grouped according to the angle ranges, with each group being evaluated based on its performance.
Among these groups, the one with angles ranging from $0^{\circ}$ to $90^{\circ}$ demonstrated the most promising results. The area under the receiver operating characteristic (AUROC) for this group was 0.68 for 2D views and an impressive 0.975 for 3D views. This highlights that the model performed significantly better in a three-dimensional context for this particular angle range.

On the other hand, the group with angles spanning from $90^{\circ}$ to $180^{\circ}$ was the lowest-performing group. The AUROC for 2D views was only 0.46, and for 3D views, it was slightly better at 0.61. This group's lower performance underscores the challenges faced when dealing with wider angles.

Following these initial experiments, we proceeded to explore datasets that featured a higher granularity of angles. Similar to the initial tests, the group with angles between $45^{\circ}$ and $90^{\circ}$ showed the best performance, with an AUROC of 0.9 for 2D views and a nearly perfect 0.99 for 3D views. 

Interestingly, several other angle groups---including those ranging from $0^{\circ}$ to $45^{\circ}$, $225^{\circ}$ to $270^{\circ}$, $270^{\circ}$ to $315^{\circ}$, and $315^{\circ}$ to $360^{\circ}$---showed nearly equivalent levels of performance for the 3D views. This finding suggests that our model maintains a high level of accuracy across various angle ranges when dealing with 3D data.

However, a decrease in performance was observed for some angle ranges. In particular, the group with angles from $180^{\circ}$ to $225^{\circ}$ had an AUROC of 0.64 for 2D views and 0.72 for 3D views, making it the worst-performing group in this set of experiments.

\begin{figure}[h]
    \centering
    \includegraphics[scale=.185,keepaspectratio]{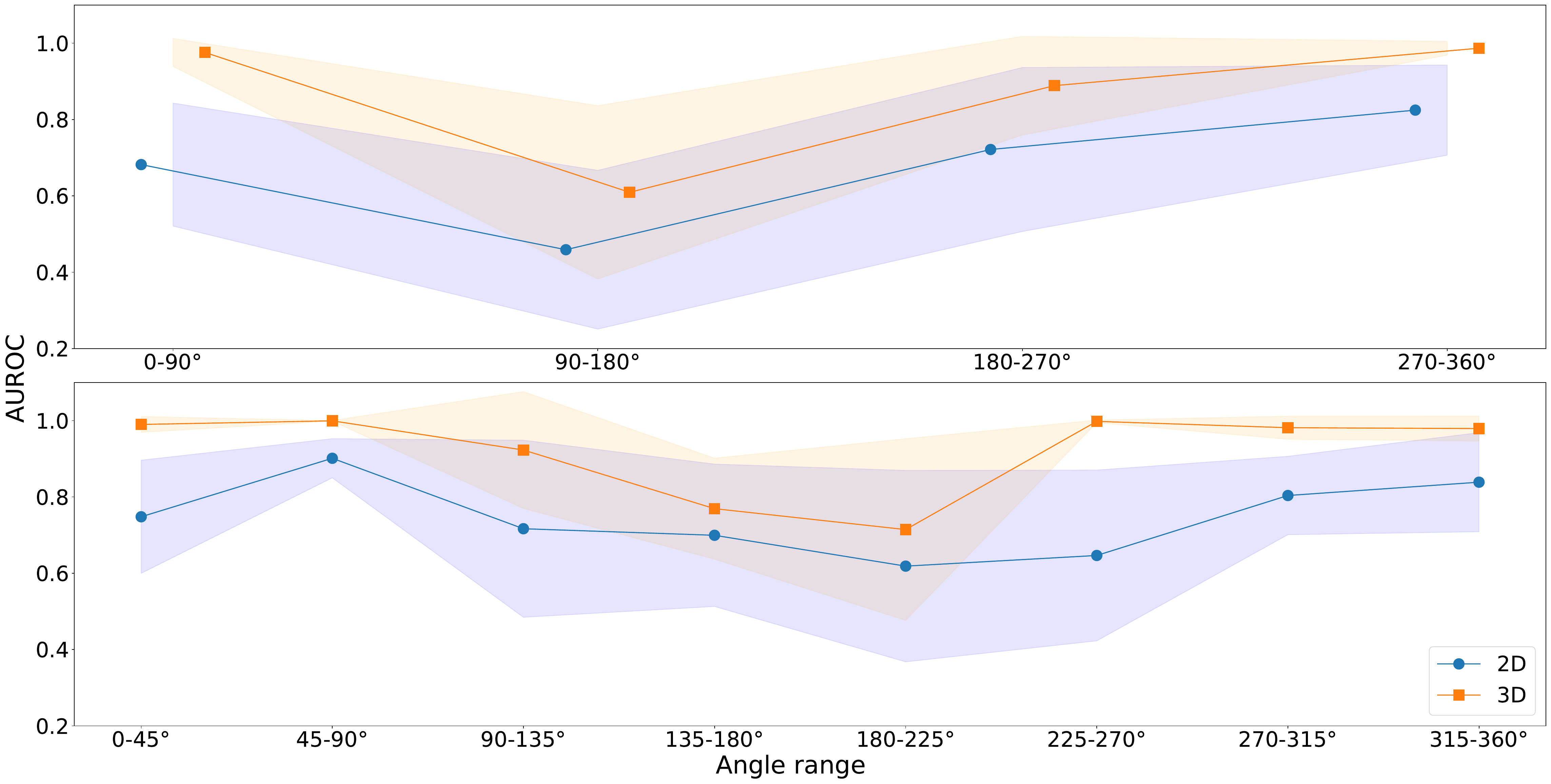}
    \caption{Average AUROC for every angle group for both angle granularities. The runs with low angle granularity are shown on the figure to the left, while the low granularity group is shown to the right one.}
    \label{fig:angle-groups-results}
\end{figure}

For a more comprehensive understanding of these findings, refer to Figure \ref{fig:angle-groups-results}, which visualizes the results of both experiments, providing a detailed comparison of the AUROC values across all angle groups and both 2D and 3D contexts.

\begin{figure}[H]
    \centering
    \includegraphics[scale=.185]{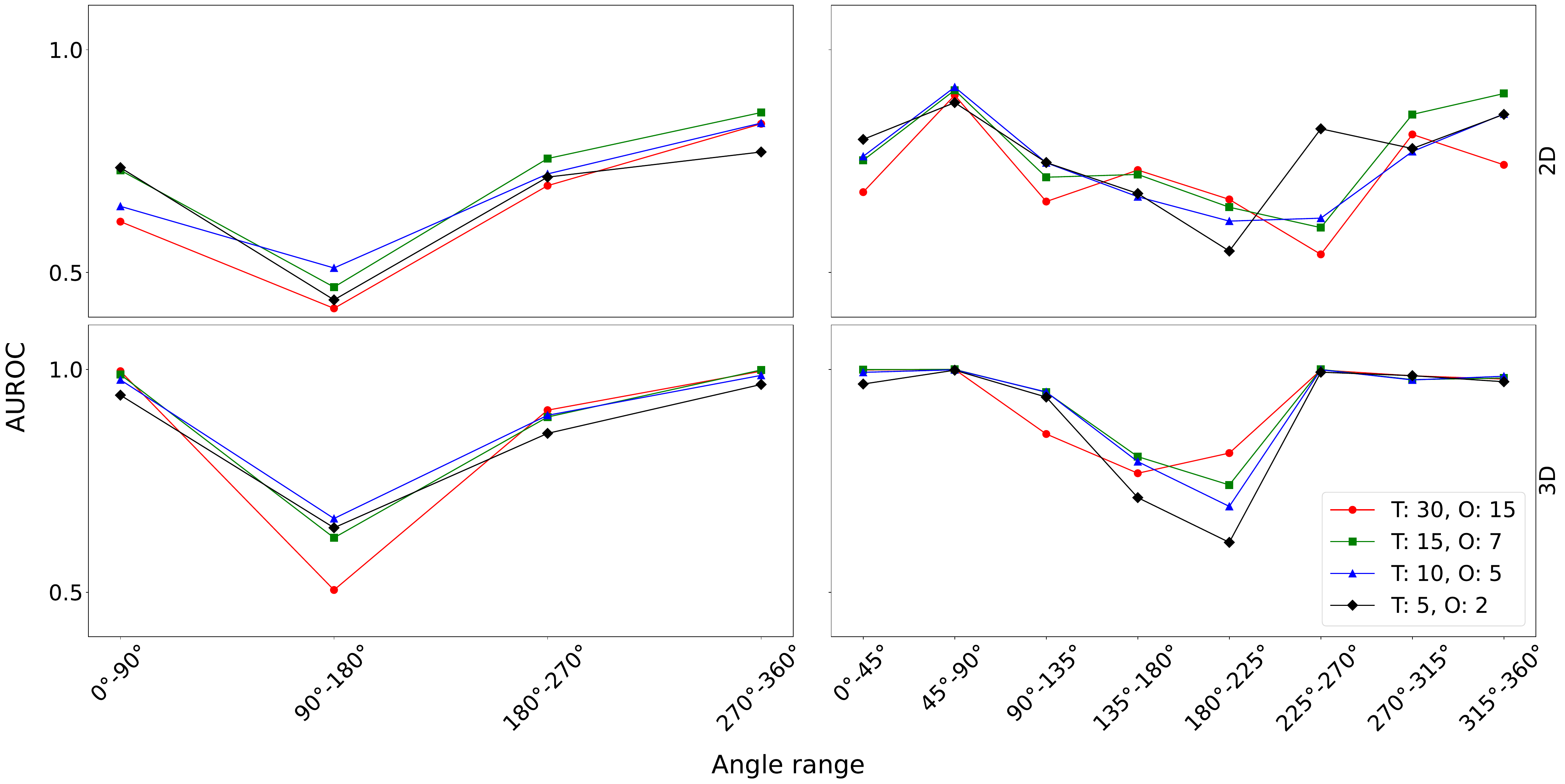}
    \caption{Average AUROC across angle groups for different sample lengths and overlap.}
    \label{fig:timestep-results}
\end{figure}

In Fig.\ref{fig:timestep-results}, the effect of sample length is shown across different angle groups. Generally, we can see a trend favoring longer samples for both 2D and 3D across all angle groups, although for 2D data there exists a point of diminishing returns, after which performance start to degrade for longer samples. For both 2D and 3D views the sharp drop in performance is observed for angles between 90\textdegree and 225\textdegree.
\section{Conclusions and discussion}
The results indicate that spatiotemporal graph attention networks are capable of distinguishing between different gait patterns in dogs. The ability of the classifier to correctly distinguish between the classes are highly dependent both on the length of samples presented to the network as well as the view of the object for which the recordings were obtained. The results confirm the common assumptions that the best performance can be achieved when the recording is taken from such an angle that affected joints are easily visible. As the camera moves into position where the joints of interest are mostly hidden the performance degrades and the prediction results start to be much less stable in terms of variability in comparison to the angles that clearly show dysfunctional joints.
Comparing approaches using points described in 2D or 3d coordinates, the 3D view provides clear advantage, showing significant performance increase across all experiments. This information is crucial for obtaining real life videos, as inclusion of depth camera or performing pose uplifting as a pre-processing step can be highly beneficial for the final model performance.
The relationship between length of the sample and final performance of the classifier is complex. Common assumption is that extending the length of a single sample will be beneficial holds true, but only when the samples are obtained from angles where the joint of interest is easily visible. For the opposite cases, the benefit of longer samples seem to diminish, possibly indicating that in those cases extending the sample length only increases the amount of irrelevant information provided to the model.

Therefore, three main conclusions emerge from the obtained results:
\begin{itemize}
    \item The visibility of the joint of interest is crucial for the performance of the model
    \item Obtaining 3D coordinates for observed points can provide significant increase in classifier's performance
    \item The training usually benefits from increasing the number of time steps in a single sample, although only when dealing with properly recorded videos
\end{itemize}

\section{Limitations and future work}

Authors acknowledge the limitations of the presented study. Firstly, the used model's architecture and hyperparameters can be optimized, possibly leading to improved performance. Secondly, the used model of the dog in the simulation environment is limited - it was used purely as baseline to provide proof of concept.

In the future, the work will be extended by incorporating real life data, both for enriching the gait model used in simulation and validate the performance of the trained model. Also, further exploration of methods for handling the missing joints in the graph would be a valuable addition, as they have shown great results in other works using graph neural networks \cite{rossi2021featureprop,you2020grape}.

\section*{Data availability}

The code is available at \url{https://github.com/szmazurek/Dog_gait_Cyfrovet} (classifier) and \url{https://github.com/Stsh4lson/cyfrovet-simulator} (simulator).

\section*{Acknowledgments}

This research was supported in part by the PLGrid infrastructure grant plglaoisi23 on the Athena computer cluster (https://docs.cyfronet.pl/display/~plgpawlik/Athena).

\bibliography{sample}
\section*{Appendix}

In tables \ref{tab3:average_low_gran} \ref{tab4:average_high_gran}, \ref{tab5:average_per_timestep_low} and \ref{tab6:average_per_timestep_high} the exact results for the experiments presented in Fig. \ref{fig:angle-groups-results} and \ref{fig:timestep-results} are shown.

\begin{table}[h]
\centering
\begin{tabular}{ll|l|l|l|l}
\multicolumn{2}{l|}{Gorup [\textdegree]}                 & 0-90                         & 90-180                       & 180-270                      & 270-360                      \\ \hline
\multicolumn{1}{l|}{\multirow{2}{*}{AUROC}} & 2D & .68 \textpm .16 & .46 \textpm .21 & .72 \textpm .21 & .82 \textpm .12 \\ \cline{2-6} 
\multicolumn{1}{l|}{}                       & 3D & .98 \textpm .04 & .61 \textpm .23 & .89 \textpm .13 & .99 \textpm .02
\end{tabular}
\caption{Average AUROC across all runs for low granularity angle groups.}
\label{tab3:average_low_gran}
\end{table}

\begin{table}[h]
\centering
\begin{tabular}{ll|p{.3in}|p{.33in}|p{.4in}|p{.465in}|p{.465in}|p{.465in}|p{.465in}|p{.465in}}
\multicolumn{2}{l|}{Group [\textdegree]}                 & 0-45                         & 45-90                         & 90-135                        & 135-180                      & 180-225                      & 225-270                       & 270-315                      & 315-360                       \\ \hline
\multicolumn{1}{l|}{\multirow{2}{*}{AUROC}} & 2D & .75 \textpm .15 & .9 \textpm .05   & .72 \textpm .23  & .7 \textpm .19  & .62 \textpm .25 & .65 \textpm .22  & .8 \textpm .1   & .84 \textpm .13  \\ \cline{2-10} 
\multicolumn{1}{l|}{}                       & 3D & .99 \textpm .02 & .99 \textpm .001 & .92 \textpm .15 & .77 \textpm .13 & .71 \textpm .24 & .99 \textpm .003 & .98 \textpm .03 & .98 \textpm .03
\end{tabular}
\caption{Average AUROC across all runs for high granularity angle groups.}
\label{tab4:average_high_gran}
\end{table}

\begin{table}[h]
\centering
\begin{tabular}{ll|l|l|l|l}
\multicolumn{2}{l|}{\multirow{2}{*}{\backslashbox{Group [\textdegree]}{T\&O}}} & \multirow{2}{*}{T: 30 O: 15}   & \multirow{2}{*}{T: 15 O: 7}   & \multirow{2}{*}{T: 10 O: 5}   & \multirow{2}{*}{T: 5 O:2}     \\
\multicolumn{2}{l|}{}                               &                                &                               &                               &                               \\ \hline
\multicolumn{1}{l|}{\multirow{2}{*}{0-90}}     & 2D & .614 \textpm .2   & .73 \textpm .17  & .65 \textpm .13  & .74  \textpm .08 \\ \cline{2-6} 
\multicolumn{1}{l|}{}                          & 3D & .99 \textpm .005  & .99 \textpm .1   & .98  \textpm .03 & .94 \textpm .05  \\ \hline
\multicolumn{1}{l|}{\multirow{2}{*}{90-180}}   & 2D & .42  \textpm .24  & .47 \textpm .24  & .51 \textpm .16  & .44  \textpm .17 \\ \cline{2-6} 
\multicolumn{1}{l|}{}                          & 3D & .51 \textpm .26  & .62 \textpm .27  & .67  \textpm .17 & .65 \textpm .16  \\ \hline
\multicolumn{1}{l|}{\multirow{2}{*}{180-270}}  & 2D & .7  \textpm .2    & .76 \textpm .22  & .72  \textpm .21 & .71 \textpm .22  \\ \cline{2-6} 
\multicolumn{1}{l|}{}                          & 3D & .91 \textpm .12   & .89 \textpm .13  & .9  \textpm .13  & .86 \textpm .14  \\ \hline
\multicolumn{1}{l|}{\multirow{2}{*}{270-360}}  & 2D & .83  \textpm .15 & .86 \textpm .09  & .84  \textpm .07 & .77 \textpm .12  \\ \cline{2-6} 
\multicolumn{1}{l|}{}                          & 3D & .99 \textpm .004  & .99 \textpm .001 & .99  \textpm .01 & .97 \textpm .02 
\end{tabular}
\caption{Average AUROC for different timestep and overlap of the samples for low granularity angle groups (T - timestep, O - overlap).}
\label{tab5:average_per_timestep_low}
\end{table}

\begin{table}[hbt]
\centering
\begin{tabular}{ll|l|l|l|l}
\multicolumn{2}{l|}{\multirow{2}{*}{\backslashbox{Group [\textdegree]}{T\&O}}} & \multirow{2}{*}{T: 30 O: 15} & \multirow{2}{*}{T: 15 O: 7}  & \multirow{2}{*}{T: 10 O: 5}  & \multirow{2}{*}{T: 5 O:2}    \\
\multicolumn{2}{l|}{}                               &                              &                              &                              &                              \\ \hline
\multicolumn{1}{l|}{\multirow{2}{*}{0-45}}     & 2D & .68 \textpm .2    & .75  \textpm .13  & .76  \textpm .13  & .8  \textpm .08   \\ \cline{2-6} 
\multicolumn{1}{l|}{}                          & 3D & .99  \textpm .002 & .99  \textpm .001 & .99  \textpm .006 & .97  \textpm 0.03 \\ \hline
\multicolumn{1}{l|}{\multirow{2}{*}{45-90}}    & 2D & .9  \textpm .04   & .91  \textpm .05  & .92  \textpm .06  & .88  \textpm .05  \\ \cline{2-6} 
\multicolumn{1}{l|}{}                          & 3D & 1  \textpm .0001  & 1  \textpm .0001  & .99  \textpm .001 & .99  \textpm .001 \\ \hline
\multicolumn{1}{l|}{\multirow{2}{*}{90-135}}   & 2D & .66  \textpm .24  & .71  \textpm .26  & .75  \textpm .24  & .75  \textpm .17  \\ \cline{2-6} 
\multicolumn{1}{l|}{}                          & 3D & .85  \textpm .27  & .95  \textpm .09  & .95  \textpm .061 & .94  \textpm .07  \\ \hline
\multicolumn{1}{l|}{\multirow{2}{*}{135-180}}  & 2D & .73  \textpm .21  & .72  \textpm .15  & .67  \textpm .21  & .68  \textpm .17  \\ \cline{2-6} 
\multicolumn{1}{l|}{}                          & 3D & .77  \textpm .14  & .8  \textpm .14   & .8  \textpm .13   & .71  \textpm .11  \\ \hline
\multicolumn{1}{l|}{\multirow{2}{*}{180-225}}  & 2D & .66  \textpm .19  & .65  \textpm .29  & .62  \textpm .26  & .55  \textpm .24  \\ \cline{2-6} 
\multicolumn{1}{l|}{}                          & 3D & .81  \textpm .18  & .74  \textpm .26  & .69  \textpm .26  & .61  \textpm .2   \\ \hline
\multicolumn{1}{l|}{\multirow{2}{*}{225-270}}  & 2D & .54  \textpm .18  & .6  \textpm .21   & .62  \textpm .23  & .82  \textpm .16  \\ \cline{2-6} 
\multicolumn{1}{l|}{}                          & 3D & .99  \textpm .002 & 1  \textpm .0001  & .99  \textpm .001 & .99  \textpm .004 \\ \hline
\multicolumn{1}{l|}{\multirow{2}{*}{270-315}}  & 2D & .81  \textpm .13  & .86  \textpm .08  & .77  \textpm .1   & .78  \textpm .08  \\ \cline{2-6} 
\multicolumn{1}{l|}{}                          & 3D & .98  \textpm .03  & .98  \textpm .05  & .98  \textpm .03  & .97  \textpm .01  \\ \hline
\multicolumn{1}{l|}{\multirow{2}{*}{315-360}}  & 2D & .74  \textpm .15  & .91  \textpm .11  & .86  \textpm .12  & .87  \textpm .08  \\ \cline{2-6} 
\multicolumn{1}{l|}{}                          & 3D & .98  \textpm .04  & .98  \textpm .04  & .95  \textpm .02  & .97  \textpm .02 
\end{tabular}
\caption{Average AUROC for different timestep and overlap of the samples for high granularity angle groups (T - timestep, O - overlap).}
\label{tab6:average_per_timestep_high}
\end{table}

\end{document}